\documentclass[journal]{IEEEtran}

\ifCLASSINFOpdf
\else
   \usepackage[dvips]{graphicx}
\fi
\usepackage{url}

\usepackage{algorithm}
\usepackage{algpseudocode}
\usepackage{mathtools}
\usepackage{amsmath}
\usepackage{amssymb}
\usepackage{subcaption}
\usepackage{graphicx}
\usepackage[flushleft]{threeparttable}
\usepackage{booktabs}
\usepackage{multirow}
\usepackage{marvosym}
\usepackage{color}
\usepackage{hyperref}

\begin{document}

\title{Robust Spatiotemporal Forecasting Using Adaptive Deep-Unfolded Variational Mode Decomposition}

\author{Osama Ahmad, Lukas Wesemann, Fabian Waschkowski, and Zubair Khalid \thanks{Osama Ahmad and Zubair Khalid are with School of Science and Engineering, Lahore University of Management Sciences, Lahore, Pakistan. Lukas Wesemann, Fabian Waschkowski, and Zubair Khalid are with the Maincode, Melbourne, Australia (email: osama\_ahmad@lums.edu.pk, lukas@maincode.com, fabian@maincode.com, zubair.khalid@lums.edu.pk). 
To facilitate reproducibility, we have made our code public on \href{https://github.com/OsamaAhmad369/MAGN}{GitHub}.
}}

\maketitle
\begin{abstract}
Accurate spatiotemporal forecasting is critical for numerous complex systems but remains challenging due to complex volatility patterns and spectral entanglement in conventional graph neural networks (GNNs). While decomposition-integrated approaches like variational mode graph convolutional network (VMGCN) improve accuracy through signal decomposition, they suffer from computational inefficiency and manual hyperparameter tuning. To address these limitations, we propose the mode adaptive graph network (MAGN) that transforms iterative variational mode decomposition (VMD) into a trainable neural module. Our key innovations include (1) an unfolded VMD (UVMD) module that replaces iterative optimization with a fixed-depth network to reduce the decomposition time (by 250$\times$ for the LargeST benchmark), and (2) mode-specific learnable bandwidth constraints ($\alpha_k$) adapt spatial heterogeneity and eliminate manual tuning while preventing spectral overlap. Evaluated on the LargeST benchmark (6,902 sensors, 241M observations), MAGN achieves an 85-95\% reduction in the prediction error over VMGCN and outperforms state-of-the-art baselines.
\end{abstract}


\IEEEpeerreviewmaketitle

\section{Introduction}
\label{sec:intro}
Accurate spatiotemporal forecasting is a foundational task for understanding and managing complex systems characterized by interconnected entities, such as transportation networks, environmental monitoring grids, and financial markets. A prime example is accurate spatiotemporal traffic forecasting, which is fundamental to intelligent transportation systems for enabling route optimization~\cite{li2018diffusion}, congestion mitigation~\cite{ahmad2024variational} and emission reduction~\cite{wang2023urban}. The inherent non-stationarity of traffic patterns, characterized by volatility from events, weather, and behavioral dynamics~\cite{liu2024largest}, poses significant challenges to prediction accuracy. Graph neural networks (GNNs) have emerged as powerful tools for modeling road networks as topological graphs \cite{li2018diffusion}, with attention-based variants like attention based spatial-temporal graph convolutional network (ASTGCN) \cite{guo2019attention} enhancing relational modeling through learnable spatiotemporal correlations. Despite these advances, GNNs suffer from spectral entanglement, that is, fail to resolve low-frequency trends (e.g., daily commutes) from high-frequency fluctuations (e.g., accident-induced congestion). This leads to error propagation in long-horizon forecasts~\cite{yu2018spatio}.

To address this challenge, decomposition-integrated GNNs have gained traction. The variational mode graph convolutional network (VMGCN) \cite{ahmad2024variational} has used variational mode decomposition (VMD) \cite{dragomiretskiy2014variational} to decompose signals into $K$ band-limited intrinsic mode functions (IMFs) before processing components through attention-augmented graph convolutional networks (GCNs). While VMGCN demonstrates significant error reduction over conventional GNNs, it is limited by two main challenges. First, its iterative VMD implementation is computationally expensive~(around 102 hours for Greater Los Angeles (3,834 sensors) in the LargeST benchmark \cite{liu2024largest}). Second, it ignores spatial heterogeneity between nodes, since parameters like mode count $K$ and bandwidth constraint $\alpha$ require manual tuning via reconstruction-loss minimization. 

Deep unfolding bridges this gap by transforming iterative algorithms into trainable neural modules \cite{monga2021algorithm}. This paradigm has been employed in various applications, including the removal of clouds from geosatellite images~\cite{imran2022deep}, power allocation in wireless networks~\cite{chowdhury2021unfolding}, image restoration~\cite{mou2022deep}, speech enhancement~\cite{hershey2014deep}, sparse coding~\cite{solomon2019deep} and traffic network imputation~\cite{deng2023graph} among many other works. For example, deep unfolding has enabled 100$\times$ speedups in applications like traffic data imputation \cite{deng2023graph} and  ultrasonic signal processing \cite{fotiadou2022deep}. Yet, no prior work has used unfolding methods for VMD, despite its proven efficacy in disentangling complex multi-scale temporal patterns. In this context, we propose the mode adaptive graph network (MAGN) that transforms iterative VMD into a trainable neural module by addressing the following research questions: 
\begin{enumerate}
    \item How can deep unfolding eliminate computational bottleneck of VMD while preserving interpretability in large-scale spatiotemporal systems?
    \item Do learnable mode-specific bandwidth constraints ($\alpha_k$) outperform fixed $\alpha$ for handling heterogeneous volatility patterns in the spatiotemporal data?
\end{enumerate}
While addressing these questions, we organize the rest of the paper as follows. Section~\ref{sec:preliminaries} provides the mathematical background on graph networks, Variational Mode Decomposition, and the ASTGCN architecture. Section~\ref{sec:magn} details our proposed mode adaptive graph network (MAGN) by introducing the unfolded VMD and integrating it into the forecasting pipeline. Section~\ref{sec:results} presents our experimental setup, results, and a comprehensive analysis of the performance, efficiency, and key components of MAGN. Finally, Section~\ref{sec:conclusion} concludes the paper.

\section{Mathematical Formulation}
\label{sec:preliminaries}
\subsection{Mathematical Preliminaries}
A graph network is constructed with distinct nodes ($N$), and the relationships between them determine the structure of the network. A directed weighted graph is denoted as $\mathcal{G}=(V, \boldsymbol{A})$, where $V$ represents the set of nodes with $\lvert V \rvert=N$ and $\boldsymbol{A} \in \mathbb{R}^{N \times N} $ is the static weighted adjacency matrix. The time series features in a graph network are defined as $\boldsymbol{\mathcal{X}}=[\boldsymbol{X}_1,\boldsymbol{X}_2,\dots,\boldsymbol{X}_N]^{\prime} \in \mathbb{R}^{N \times T}$, where $\boldsymbol{X}_n \in \mathbb{R}^{T \times 1}$ is the 1-D signal for node $n$ and $T$ is the length of the signal.  The Laplacian matrix is defined as $\boldsymbol{\Tilde{L}}=\boldsymbol{D}-\boldsymbol{A}$, where $\boldsymbol{D} \in \mathbb{R}^{N \times N}$  is the diagonal matrix containing the degree of each node $\boldsymbol{D}_{ii}=\sum_j  \boldsymbol{A}_{(i,j)}$. The normalized Laplacian matrix is expressed as $\boldsymbol{L}=\boldsymbol{I}_N-\boldsymbol{D}^{-\frac{1}{2}}\boldsymbol{A}\boldsymbol{D}^{-\frac{1}{2}}$, where $\boldsymbol{I}_N$ is an identity matrix of order $N$. In a two-stage architecture, the first neural network ($\mathcal{D}_\psi$) parameterized by $\psi$ that learns to decompose each feature vector into $K$ segments~(mode-expanded representation), which preserve the temporal dynamics, that is,
$\mathcal{D}_\psi(\boldsymbol{\mathcal{X}}) = \boldsymbol{\mathcal{U}}=\left[ \boldsymbol{U_1}, \boldsymbol{U_2}, \dots, \boldsymbol{U_N} \right]^{\prime} \in \mathbb{R}^{N \times T \times K}$,    
where $\boldsymbol{\mathcal{U}}$ is a tensor of mode features. The optional $d$ features can be added to $\boldsymbol{\mathcal{U}}$ to obtain $\boldsymbol{\mathcal{Z}}=\left[ \boldsymbol{Z_1}, \boldsymbol{Z_2}, \dots, \boldsymbol{Z_N} \right]^{\prime} \in \mathbb{R}^{N \times T \times (K+d)}$, which serves as input to the second network to learn the mapping function $h$ from the historical observations data from the steps $T_w$ to predict future features for the steps $T_w^{\prime}$, that is, $[\boldsymbol{\mathcal{Z}}_{\scriptscriptstyle{1,(t-T_w+1:t)}},\dots,\boldsymbol{\mathcal{Z}}_{\scriptscriptstyle{N,(t-T_w+1:t)}};\mathcal{G}] \stackrel{h}{\rightarrow} [\boldsymbol{X}_{\scriptscriptstyle{1,(t+1:t+T_w^{\prime})}},\dots,\boldsymbol{X}_{\scriptscriptstyle{N,(t+1:t+T_w^{\prime})}}]$.

\subsection{Variational Mode Decomposition~(VMD) Driven ASTGCN}
\label{subsec:vmd_astgcn}

VMD extracts intrinsic mode functions (IMFs) by solving the constrained optimization problem~\cite{dragomiretskiy2014variational}:

\begin{equation}
\label{eq:mode}
   \hat{u}_k^{(n+1)}=\frac{\hat{f}(\omega)-\sum\limits_{i< k} \hat{u_i}^{n+1}(\omega)-\sum\limits_{i > k}\hat{u_i}^{n}(\omega)+\frac{\hat{\lambda}^n(\omega)}{2}}{1+2\alpha(\omega-\omega^n_k)^2},
\end{equation}
where $\hat{\lambda}$ is the Lagrangian multiplier, $\alpha$ the bandwidth constraint, and $\hat{f}(\omega)$ the discrete Fourier transform~(DFT) of the signal with mirrored boundaries. Center frequencies update as
\begin{equation}
\label{eq:omega_k_eq}
\omega_k^{(n+1)} = \frac{\sum_{\omega=T}^{2T} \omega |\hat{u}_k^{n+1}(\omega)|^2}{\sum_{\omega=T}^{2T} |\hat{u}_k^{n+1}(\omega)|^2}.
\end{equation}
\noindent Compared to EMD-based methods \cite{zhao2023hybrid}, VMD avoids mode mixing through its constrained optimization. Despite its theoretical advantages, iterative optimization is prohibitively expensive for large-scale spatiotemporal systems as its computational cost is $\mathcal{O}(N\mathcal{N}KT)$, where $\mathcal{N}$ is the iteration count.

The ASTGCN backbone~\cite{guo2019attention} employs dual attention mechanisms to capture node relationships and temporal correlations as
$\mathbf{S} = \mathbf{V}_s \sigma\left(\boldsymbol{\mathcal{Z}}\mathbf{W}_1\mathbf{W}_2(\mathbf{W}_3\boldsymbol{\mathcal{Z}})^T + \mathbf{b}_s\right)$ and
$\mathbf{E} = \mathbf{V}_e \sigma\left((\boldsymbol{\mathcal{Z}}^T\mathbf{V}_1)\mathbf{V}_2(\mathbf{V}_3\boldsymbol{\mathcal{Z}}) + \mathbf{b}_e\right)$, respectively. Here $\mathbf{V}_s, \mathbf{b}_s \in \mathbb{R}^{N \times N}$, $\mathbf{W}_1 \in \mathbb{R}^{T_w}$, $\mathbf{W}_2 \in \mathbb{R}^{(K+d) \times T_w}$, and $\mathbf{W}_3 \in \mathbb{R}^{K+d}$ are trainable parameters for spatial attention, while $\mathbf{V}_e, \mathbf{b}_e \in \mathbb{R}^{T_w \times T_w}$, $\mathbf{V}_1 \in \mathbb{R}^{N}$, $\mathbf{V}_2 \in \mathbb{R}^{N \times (K+d)}$, and $\mathbf{V}_3 \in \mathbb{R}^{(K+d)}$ are parameters for temporal attention. The input tensor $\boldsymbol{\mathcal{Z}} \in \mathbb{R}^{N \times (K+d) \times T_w}$ contains $K$ VMD modes and $d$ auxiliary features, and $\sigma$ denotes the element-wise sigmoid activation. The spectral convolution implements a Chebyshev polynomial approximation~\cite{kipf2016semi} given by
\begin{equation}
g_\theta(\boldsymbol{L}) = \sum_{m=0}^{M-1} \theta_m T_m(\hat{\boldsymbol{L}}) \odot \mathbf{S}',\quad \hat{\boldsymbol{L}} = \tfrac{2}{\lambda_{\text{max}}}\boldsymbol{L} - \mathbf{I}_n,
\end{equation}
where $\theta_m$ are learnable coefficients, $T_m$ is the $m$-th order Chebyshev polynomial, $\lambda_{\rm {max}}$ is the maximum value of $\boldsymbol{L}$, and $\odot$ denotes Hadamard product with spatial attention $\mathbf{S}'$ (normalized attention weights obtained through softmax normalization of $\mathbf{S}$). 
\begin{figure*}[!t]
\centering
\vspace{-1mm}
\includegraphics[width=0.8\linewidth]{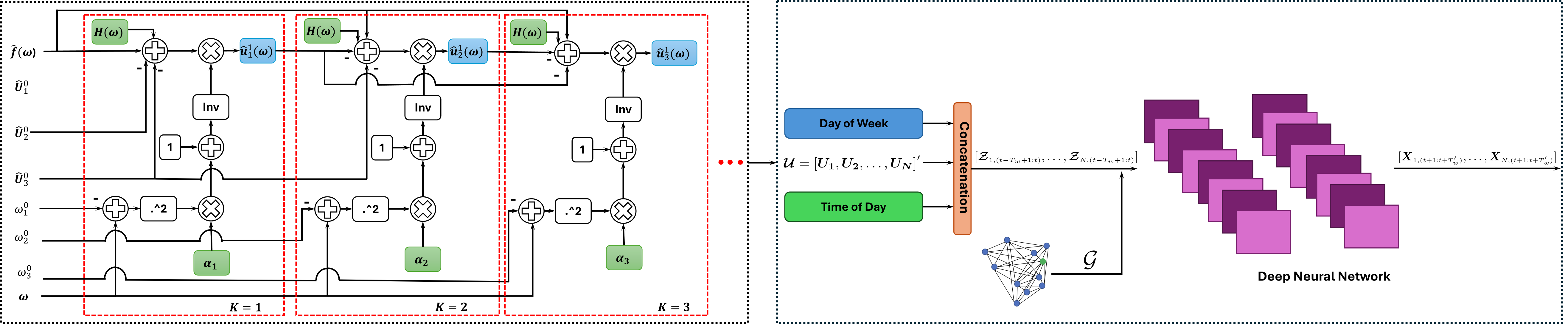}
\caption{
Proposed Two-stage MAGN architecture. Stage 1: The Unfolded VMD network is trained to decompose the input signal into multiscale mode features by minimizing the reconstruction loss. Stage 2: Mode features are concatenated with additional features and fed into a spatiotemporal ASTGCN trained to predict future states by minimizing the prediction loss MAE.}
\vspace{-1mm}
\label{fig:unfold_VMD}
\end{figure*}

\begin{table*}[!t]
\centering
\caption{Comparison of performance metrics MAE, MAPE, and RMSE between different baselines and our model on horizons $3$, $6$, $12$, and average. The average results are computed using the mean from the horizon of $1$ to $12$. The number of parameters (param) is described in K (kilo), $10^3$, and M (million), $10^6$, and the best performance metrics are highlighted in red bold numbers. * numbers are taken from~\cite{ahmad2025robust}. In the param column, in \textcolor{red}{x}+\textcolor{blue}{y}, \textcolor{red}{x} shows the parameters of the UVMD and \textcolor{blue}{y} represents the parameters of the prediction model.  {\large \texttimes} parameters count is not publicly available. All baselines were retrained on the LargeST splits with the same normalization and evaluation protocol as MAGN.
 }
 \vspace{-1mm}
\resizebox{.93\textwidth}{!}{
\begin{threeparttable}
\begin{tabular}{c|cc|ccc|ccc|ccc|ccc}

\toprule
 Dataset & Method & Param & \multicolumn{3}{c}{Horizon 3} & \multicolumn{3}{c}{Horizon 6} & \multicolumn{3}{c}{Horizon 12} & \multicolumn{3}{c}{Average}  \\
\cmidrule(r){4-15}
& & &MAE & RMSE & MAPE& MAE & RMSE& MAPE & MAE & RMSE & MAPE& MAE & RMSE & MAPE \\
\midrule
\multirow{4}{*}{GBA}
& HL* & - & 32.57 & 48.42 & 22.78\% & 53.79 & 77.08 & 43.01\% & 92.64 & 126.22 & 92.85\% & 56.44 & 79.82 & 48.87\% \\
& LSTM* & 98K & 20.41 & 33.47 & 15.60\% & 27.50 & 43.64 & 23.25\% & 38.85 & 60.46 & 37.47\% & 27.88 & 44.23 & 24.31\% \\
& ASTGCN* & 22.30M & 21.40 & 33.61 & 17.65\% & 26.70 & 40.75 & 24.02\% & 33.64 & 51.21 & 31.15\% & 26.15 & 40.25 & 23.29\% \\
& D$^2$STGNN* & 446K & 17.20 & 28.50 & 12.22\% & 20.80 & 33.53 & 15.32\% & 25.72& 40.90 & 19.90\% & 20.71 & 33.44 & 15.23\% \\
&PatchSTG~\cite{fang2024efficient}& 3.11M&16.81 &28.71 &12.25\% & 19.68 &33.09& 14.51\%  &23.49 &39.23 &18.93\%  &19.50& 33.16 &14.64\% \\
&RPMixer~\cite{yeh2024rpmixer} &2.30M& 17.35& 28.69& 13.42\% &19.44& 32.04& 15.61\% &21.65 &36.20 &17.42\%& 19.06& 31.54 &15.09\%\\
&RAGL~\cite{wu2025regularized} &{\large \texttimes}&15.71 &27.58& 10.29\%& 18.40 &31.89 &12.23\%& 22.48 &38.39 &15.92\% &18.33& 31.65 &12.18\%\\
& VMGCN* & 22.38M &2.90 & 5.32 & 3.27\% & 6.47 & 11.62 &6.86\% & 16.42 & 26.45 & 17.55\% & 8.04 & 13.55 & 8.57\% \\
& CA-VMGCN* & 22.40M & 3.50 & 6.19 & 3.91\% & 6.59 & 11.50 & 6.89\% & 14.77 & 23.47&15.27\% & 7.77& 12.90&8.14\%\\
&MAGN (Ours) & 140.174K+22.38M & \textbf{\textcolor{red}{0.62}} & \textbf{\textcolor{red}{0.93}} & \textbf{\textcolor{red}{0.68\%}} & \textbf{\textcolor{red}{0.68}} & \textbf{\textcolor{red}{1.09}} & \textbf{\textcolor{red}{0.74\%}} &\textbf{\textcolor{red}{ 2.00}} & \textbf{\textcolor{red}{4.71}}&\textbf{\textcolor{red}{1.90\%}} & \textbf{\textcolor{red}{0.86}}& \textbf{\textcolor{red}{1.63}}&\textbf{\textcolor{red}{0.91\%}}\\
\midrule

\multirow{4}{*}{GLA} 
& HL*& - & 33.66 & 50.91 & 19.16\% & 56.88 & 83.54 & 34.85\% & 98.45 & 137.52 & 71.14\% & 56.58 & 86.19 & 38.76\% \\
& LSTM* & 98K & 20.09 & 32.41 & 11.82\% & 27.80 & 44.10 & 16.52\% & 39.61 & 61.57 & 25.63\% & 28.12 & 44.40 & 17.31\% \\
& ASTGCN*  & 59.1M & 21.11 & 32.41 & 11.82\% & 27.80 & 44.67 & 17.79\% & 39.39 & 59.31 & 28.03\% & 28.12 & 44.40 & 18.62\% \\
& D$^2$STGNN* & 284K & 19.31 & 30.07 & 11.82\% & 22.52 & 35.22 & 14.16\% & 27.46& 43.37 & 18.54\% & 22.35 & 35.11 & 14.37\% \\
&PatchSTG~\cite{fang2024efficient}&1.68M& 15.84 &26.34 &9.27\%& 19.06 &31.85& 11.30\% &23.32 &39.64 &14.60\% &18.96 &32.33 &11.44\% \\
&RPMixer~\cite{yeh2024rpmixer}&3.20M &16.49 &26.75& 9.75\%& 18.82 &30.56& 11.58\% &21.18& 35.10& 13.46\% &18.46 &30.13 &11.34\%\\
&RAGL~\cite{wu2025regularized}&{\large \texttimes} &15.06 &25.66& 8.39\%& 17.84& 30.24& 10.09\%& 21.72 &36.73& 12.98\% &17.75& 30.11& 10.20\%\\
& VMGCN* & 59.2M &3.88& 10.78 & 3.99\% & 8.27 & 22.34 & 7.85\%& 16.78 & 31.46 & 14.28\% & 9.22 & 20.69 & 8.23\%\\
& CA-VMGCN* & 59.3M & 4.37 & 8.99& 8.04\% &8.19 & 15.54 & 7.86\% &15.15& 24.08 & 12.42\%&8.85 & 15.53 & 8.76\% \\
& MAGN (Ours) & 140.173K+59.2M & \textbf{\textcolor{red}{0.67}}& \textbf{\textcolor{red}{3.16}}& \textbf{\textcolor{red}{0.62\%}} &\textbf{\textcolor{red}{0.74}} & \textbf{\textcolor{red}{4.30}} & \textbf{\textcolor{red}{0.67\%}} &\textbf{\textcolor{red}{2.73}}& \textbf{\textcolor{red}{17.71}} & \textbf{\textcolor{red}{1.89\%}}&\textbf{\textcolor{red}{1.11}} & \textbf{\textcolor{red}{6.53}} & \textbf{\textcolor{red}{0.92\%}} \\
\midrule

\multirow{4}{*}{SD} 
& HL* & - & 33.61 & 50.97 & 20.77\% & 57.80 & 84.92 & 37.73\% & 101.74 & 140.14 & 76.84\% & 60.79 & 87.40 & 41.88\% \\
& LSTM* & 98K & 19.17 & 30.75 & 11.85\% & 26.11 & 41.28 & 16.53\% & 38.06 & 59.63 & 25.07\% & 26.73 & 42.14 & 17.17\% \\
& ASTGCN* & 2.15M & 19.68 & 31.53 & 12.20\% & 24.45 & 38.89 & 15.36\% & 31.52 & 49.77 & 22.15\% & 26.07 & 38.42 & 15.63\% \\
& D$^2$STGNN* & 406K & 15.76 & 25.71 & 11.84\% & 18.81 & 30.68 & 14.39\% & 23.17 & 38.76 & 18.13\% & 18.71 & 30.77 & 13.99\% \\
&PatchSTG~\cite{fang2024efficient} &2.28M&14.53& 24.34& 9.22 \%&16.86 &28.63 &11.11\%& 20.66 &36.27& 14.72\% &16.90 &29.27 &11.23\%\\
&RPMixer~\cite{yeh2024rpmixer}&1.50M&15.12& 24.83 &9.97\% & 17.04  &28.24 & 10.98\%  &19.60 & 32.96 & 13.12\% & 16.90 & 27.97  &11.07\%\\
&RAGL~\cite{wu2025regularized} &{\large \texttimes}&13.87& 23.42& 9.01\% &16.09& 27.35& 10.63\% &19.90& 33.94& 13.35\% &16.16& 27.40& 10.62\%\\
&VMGCN*  & 2.17M&6.67 & 13.51 & 6.02\% & 11.25 & 27.96 & 10.23\% & 20.73 & 85.97 & 20.80\% & 12.23 & 39.48& 11.69\% \\
&CA-VMGCN* & 2.19M& 7.17 & 12.27& 6.08\% & 11.27 & 20.46 & 9.20\% & 18.44 & 38.97 & 15.89\% & 11.71 & 22.56 & 9.79\%\\
&MAGN (Ours) & 140.173K+2.17M&  \textbf{\textcolor{red}{0.84}}	&\textbf{\textcolor{red}{1.88}}&	\textbf{\textcolor{red}{0.84\%}}&\textbf{\textcolor{red}{0.90}}&\textbf{\textcolor{red}{2.23}}&	\textbf{\textcolor{red}{0.92\%}}&\textbf{\textcolor{red}{3.60}}&\textbf{\textcolor{red}{8.49}}&\textbf{\textcolor{red}{2.97\%}}&\textbf{\textcolor{red}{1.38}}&\textbf{\textcolor{red}{3.30}}&\textbf{\textcolor{red}{1.28\%}}\\
\hline
\end{tabular}
\end{threeparttable}
}
\label{tab:comparison}
\end{table*}

\section{Mode Adaptive Graph Network~(MAGN) Driven by Neural VMD}
\label{sec:magn}

This work proposes the mode adaptive graph network~(MAGN) architecture to address the research questions posed in Section~\ref{sec:intro} through the following key contributions:

\begin{itemize}
    \item We introduce the first neural implementation of VMD by unrolling its alternating direction method of multipliers (ADMM) iterations into a differentiable module (Fig. 1). The proposed unfolded VMD achieves significant reduction in computation cost while maintaining interpretability.
    \item The learnable parameters in an unrolling algorithm provide an ease in tuning of parameters for each signal in a spatiotemporal network and enable dynamic adaptation to local volatility patterns (e.g., highway vs. urban sensors). 
    \item Evaluated on LargeST with 6,902 sensors and 241 million observations, MAGN enables {85-95\% error reduction} over VMGCN in MAE/MAPE/RMSE, {250x speedup} in decomposition (267~minutes (mins) to 98.63~seconds (s) for LargeST), and frequency-level interpretability~(revealing {rush-hour harmonics} and {event-driven anomalies}).    
\end{itemize}

\subsection{Unfolded Variational Mode Decomposition (UVMD)}
\label{subsec:uvmd}

We transform the iterative VMD algorithm into a trainable neural module to overcome computational bottlenecks and enable parameter adaptation~(see Fig.~\ref{fig:unfold_VMD}). The unfolded mode update equation is
\begin{equation}
\label{eq:learnable_mode_eq}
\hat{u}_k^{(n+1)}(\omega) = \frac{\hat{f}(\omega) - \sum_{i<k} \hat{u}_i^{n+1}(\omega) - \sum_{i>k} \hat{u}_i^n(\omega) + \frac{{H}^n(\omega)}{2}}{1 + 2\phi(\alpha_k)(\omega - \omega_k^n)^2}
\end{equation}
where ${H}(\omega)$ denotes a learnable complex-valued Lagrangian multiplier, $\alpha_k$ is a learnable mode-specific bandwidth constraint parameter, and $\phi(\cdot) = \log(1 + e^{(\cdot)})$ is the SoftPlus function. Intuitively, this update balances the input spectrum with residual modes, while $\alpha_k$ adaptively controls the sharpness of the frequency band of each mode. The reconstruction loss enforces spectral fidelity and is given by
$\mathcal{L}_{\text{rec}} = \| \hat{f}(\omega) - \sum_{k=1}^K \hat{u}_k(\omega) \|_2$. This unrolled structure replaces iterative convergence checks with a shallow network ($\mathcal{N} = 1$--$2$ layers), reduces decomposition time and enables the adaptation of $\alpha_k$ to spatial heterogeneity. \looseness=-1 

\subsection{Architecture}
\begin{table*}[!t]
\centering
\caption{ Performance evaluation of metrics MAE, MAPE, and RMSE on SD region.}
\vspace{-1mm}
\resizebox{.95\textwidth}{!}{
\begin{tabular}{c|c|c|ccc|ccc|ccc|ccc}
\toprule
 Case &Description &Parameters&   \multicolumn{3}{c}{Horizon 3} & \multicolumn{3}{c}{Horizon 6} & \multicolumn{3}{c}{Horizon 12} & \multicolumn{3}{c}{Average}  \\
\cmidrule(r){4-15}
Scenario& &&MAE & RMSE & MAPE& MAE & RMSE& MAPE & MAE & RMSE & MAPE& MAE & RMSE & MAPE \\
\midrule
\multirow{2}{*}{Case I ($\alpha$)  K=13, $\mathcal{N}$=1}
& Shared ($\alpha$)& 140.161K+2.17M&1.61&	3.84&	1.53\%&	14.69&	23.54&	9.46\%&	28.28&	45.20&	19.75\%&	13.70&	22.70&	9.34\%\\
& Mode-specific ($\alpha_k$)& 140.173K+2.17M&0.84&1.88&0.84\%&0.90&2.23&0.92\%&3.60&8.49&2.97\%&1.38&3.30&1.28\% \\

\midrule
\multirow{4}{*}{Case II (signal length) K=13, $\mathcal{N}$=1} 
&Full signal (35040)& 140.173K+2.17M& 0.84	&1.88&	0.84\%&0.90&	2.23&	0.92\%&	3.60&	8.49&	2.97\%&	1.38&	3.30&	1.28\%\\
&1/2 signal (17520) &  70.093K+2.17M& 0.88	&2.08	&0.96\%&1.05	&2.85	&1.17\%&5.05&	13.27	&4.79\%&	1.80&4.90&	1.83\% \\
&1/4 signal (8760) &  35.053K+2.17M&0.87&	2.09&	0.98\%&	1.06&	2.91&	1.22\%&	6.37&	17.06&	5.56\%&	2.03&	5.54&	2.01\%\\
&1/8 signal (4380) &  17.533K+2.17M& 0.94&2.92&	1.01\%&	1.18&	4.48&	1.28\%&	7.01&	28.89&	5.78\%&	2.26&	9.38&	2.09\% \\
\midrule

\multirow{7}{*}{Case III (hyper-parameters)}
&$K$=3, $\mathcal{N}$=1&140.163K+2.16M&7.65&	14.51&	5.42\%&	11.80&	20.35&	8.25\%&	25.24&	69.03&	23.41\%&	13.56&	29.76&	11.04\%\\
&$K$=6, $\mathcal{N}$=1&140.166K+2.16M&0.97&2.48&0.94\%&2.33&7.69&1.86\%&10.08&16.18&7.82\%&4.00&8.72&3.11\%\\
&$K$=9, $\mathcal{N}$=1&140.169K+2.16M&0.86&1.94&0.86\%&	0.94&	2.48&	0.93\%&	5.93&	12.74&	4.17\%&	1.89&	4.52&	1.59\%\\
&$K$=13, $\mathcal{N}$=1& 140.173K+2.17M&0.84&1.88&0.84\%&0.90&2.23&0.92\%&3.60&8.49&2.97\%&1.38&3.30&1.28\% \\
&$K$=15, $\mathcal{N}$=1&140.175K+2.17M&0.85&1.72&0.92\%&0.94&1.80&1.01\%&2.47&6.34&2.18\%&1.16&2.60&1.19\% \\
&$K$=6, $\mathcal{N}$=2&210.246K+2.16M&5.35&8.84&5.22\%&9.71&15.15&8.93\%&21.14&32.36&22.01\%&11.19&17.57&10.92\% \\
&$K$=13, $\mathcal{N}$=2&210.253K+2.17M&7.42&12.47&9.53\%&13.27&18.57&14.87\%&23.62&33.30&29.22\%&13.49&19.82&16.65\% \\


\hline
\end{tabular}
}
\vspace{-2mm}
\label{tab:hyperparameter}
\end{table*}

The VMD through ADMM optimization~\cite{dragomiretskiy2014variational} is carried out using two nested loops. The outer loop is applied until all modes converge while the inner loop computes $K$ modes per outer iteration. Fig.~\ref{fig:unfold_VMD} shows the inner loop of this iterative algorithm for $\mathcal{N}=1$. This step is further divided into two steps: computation of $k^{\rm {th}}$ mode represented by $\boldsymbol{\hat{U}}_k(\omega) \in \mathbb{C}^{\mathcal{N}\times 2T\times K}$ and the update of center frequency $\omega_k \in \mathbb{R}^{\mathcal{N} \times K}$. $\hat{f}\in \mathbb{C}^{\mathcal{N}\times 2T}$ is the signal in the frequency domain; twice the length indicates that the mirror signal around the center axis of each sequence is used to avoid the boundary discontinuity. $H$ and $\alpha_k$ are learnable parameters, where $H$ is considered as a complex bias parameter $H \in \mathbb{C}^{\mathcal{N}\times 2T}$ and $\alpha_k$ is the positive real number for mode-specific bandwidth constraint. The Gauss-Seidel method is used to determine the modes for $\mathcal{N}$ iterations. Using this method, the modes converge in fewer iterations. In Fig.~\ref{fig:unfold_VMD}, three modes are used to demonstrate the evolution from $\textit{n}=0$ to $\textit{n}=1$. The parameters represented in a green block are learnable parameters, and the blue block indicates the output of each mode. In the first stage, UVMD decomposes the features into mode vectors. This network updates its parameters by minimizing the reconstruction loss using gradient descent. Followed by the decomposition, the modes serve as an input to the ASTGCN to predict future states. The mean absolute error (MAE) is used to learn the parameters of ASTGCN. It is important to note that UVMD does not perform task-specific learning or directly optimize for forecasting objectives. The module only adapts the bandwidth parameters $\alpha_k$ and Lagrange multipliers $H$ that govern the decomposition of input signals into interpretable modes. Since these parameters are global rather than sequence-specific, we ensure that the training of UVMD does not introduce label information or future values into the forecasting stage. The ASTGCN predictor subsequently learns exclusively from the decomposed modes within its training split.

\begin{figure}[!t]
    \centering
    \vspace{-2mm}
   \begin{subfigure}[b]{0.22\textwidth}
       \centering
    \includegraphics[width=\textwidth]{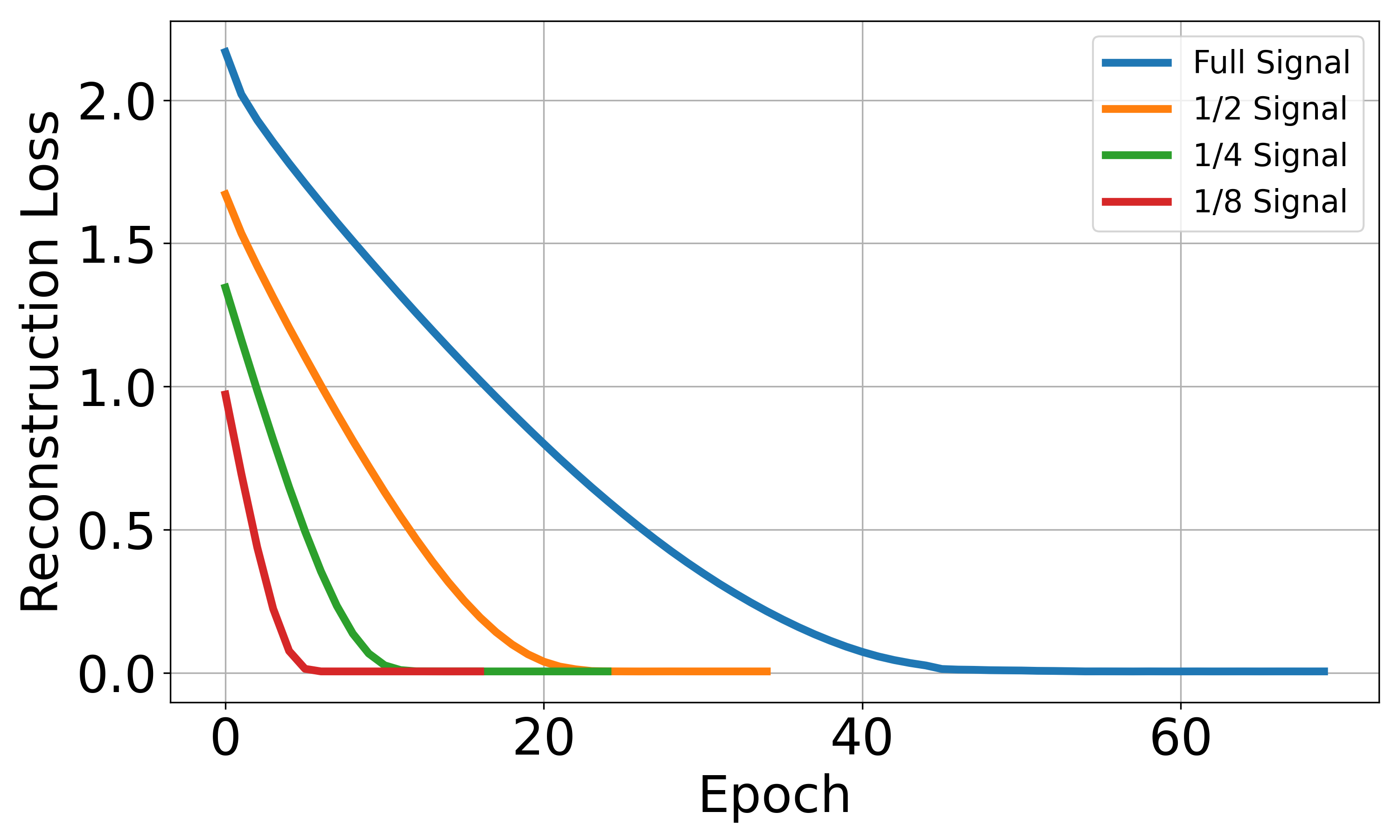}
    \vspace{-1mm}
    \label{fig:training_curve_unfold}
   \end{subfigure}
   \begin{subfigure}[b]{0.22\textwidth}
       \centering
       \vspace{-2mm}
    \includegraphics[width=\textwidth]{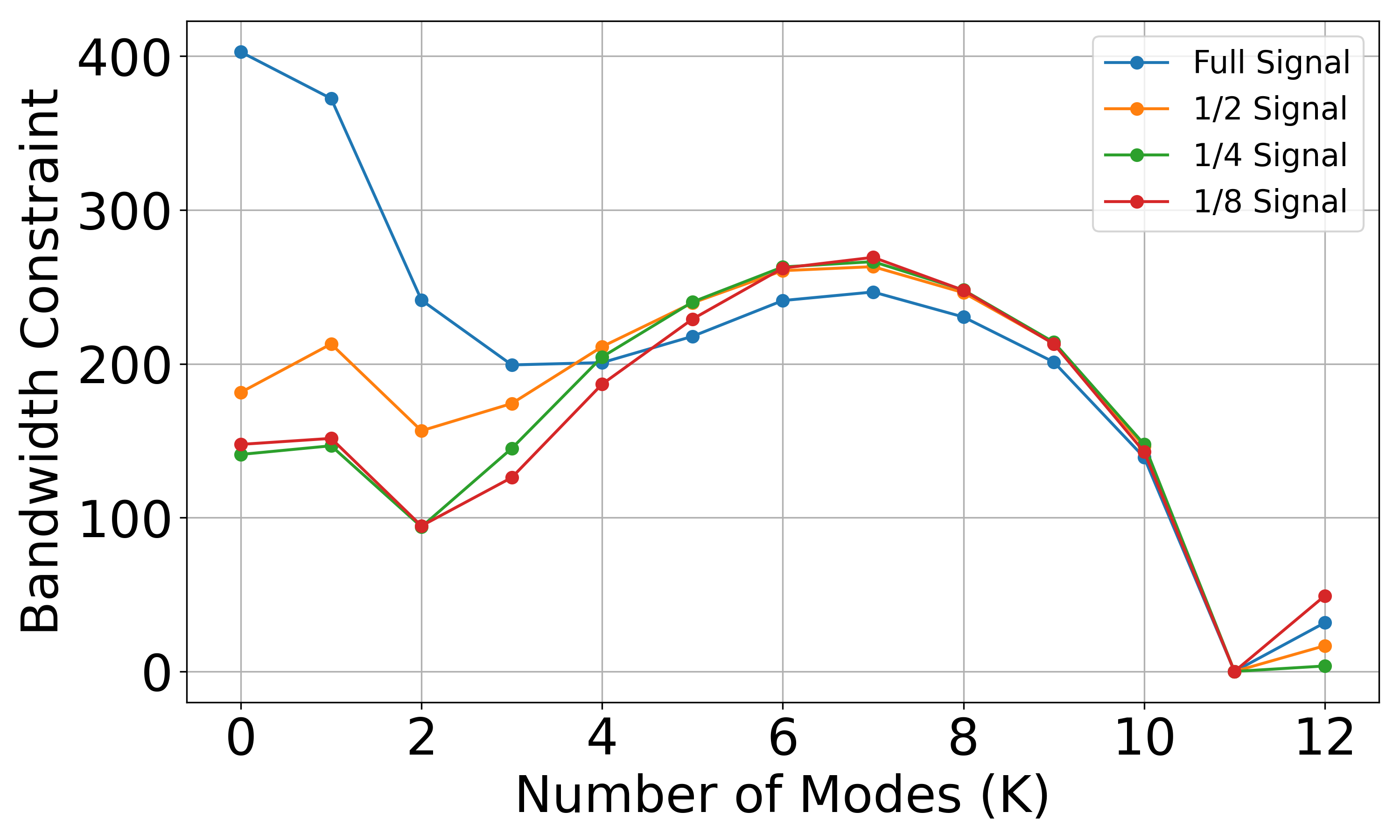}
    \vspace{-1mm}
    \label{fig:alpha_with_length}
   \end{subfigure}
   \vspace{-3mm}
   \caption{Impact assessment of different window sizes on network training~(left) and bandwidth constraint $\phi(\alpha_k)$~(right).}       \vspace{-3mm}
   \label{fig2}
\end{figure}

\section{Evaluation}
\label{sec:results}
We evaluate our model on the {LargeST benchmark} for traffic flow prediction, comprising three regions: Greater Bay Area (GBA) with $2{,}352$ nodes, Greater Los Angeles (GLA) with $3{,}834$ nodes, and San Diego (SD) with $716$ nodes. Performance is measured using MAE, mean absolute percentage error (MAPE), and root mean square error (RMSE). All experiments run on a Linux system with an Intel i9 processor, 24GB RAM, and NVIDIA 3080Ti GPU, trained on 2019 data sampled at 15-minute intervals. For the UVMD module, data is split into 70\% training, 15\% validation, and 15\% testing with batch size $1$, where $\alpha$ and $H(\omega)$ initialize to $2{,}000$ and zeros, respectively. For the forecasting module, data splits are 60\% training, 20\% validation, and 20\% testing with batch sizes of $48$ (SD) and $4$ (GLA/GBA). We use a historical window $T_w = 12$ (3 hours) to predict $T_w^{\prime} = 12$ future steps, trained with Adam optimizer for 100 epochs with early stopping.\looseness=-1
\vspace{-1mm}

\subsection{Analysis}
\begin{figure}[!t]
    \centering
    \vspace{-2mm}
   \begin{subfigure}[b]{0.22\textwidth}
       \centering
    \includegraphics[width=\textwidth]{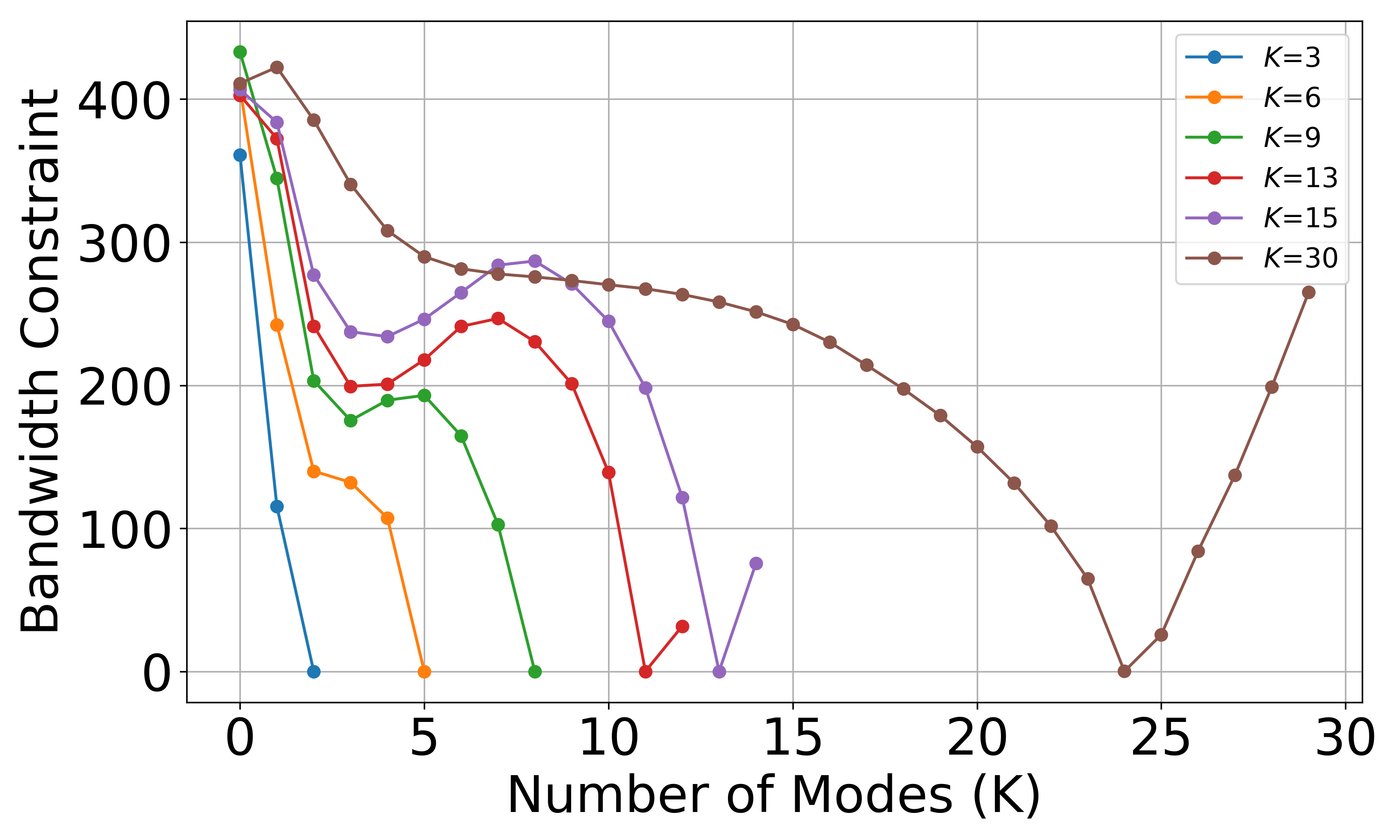}
    \vspace{-1mm}
    \label{fig:alpha}
   \end{subfigure}
   \begin{subfigure}[b]{0.22\textwidth}
       \centering
       \vspace{-2mm}
    \includegraphics[width=\textwidth]
    {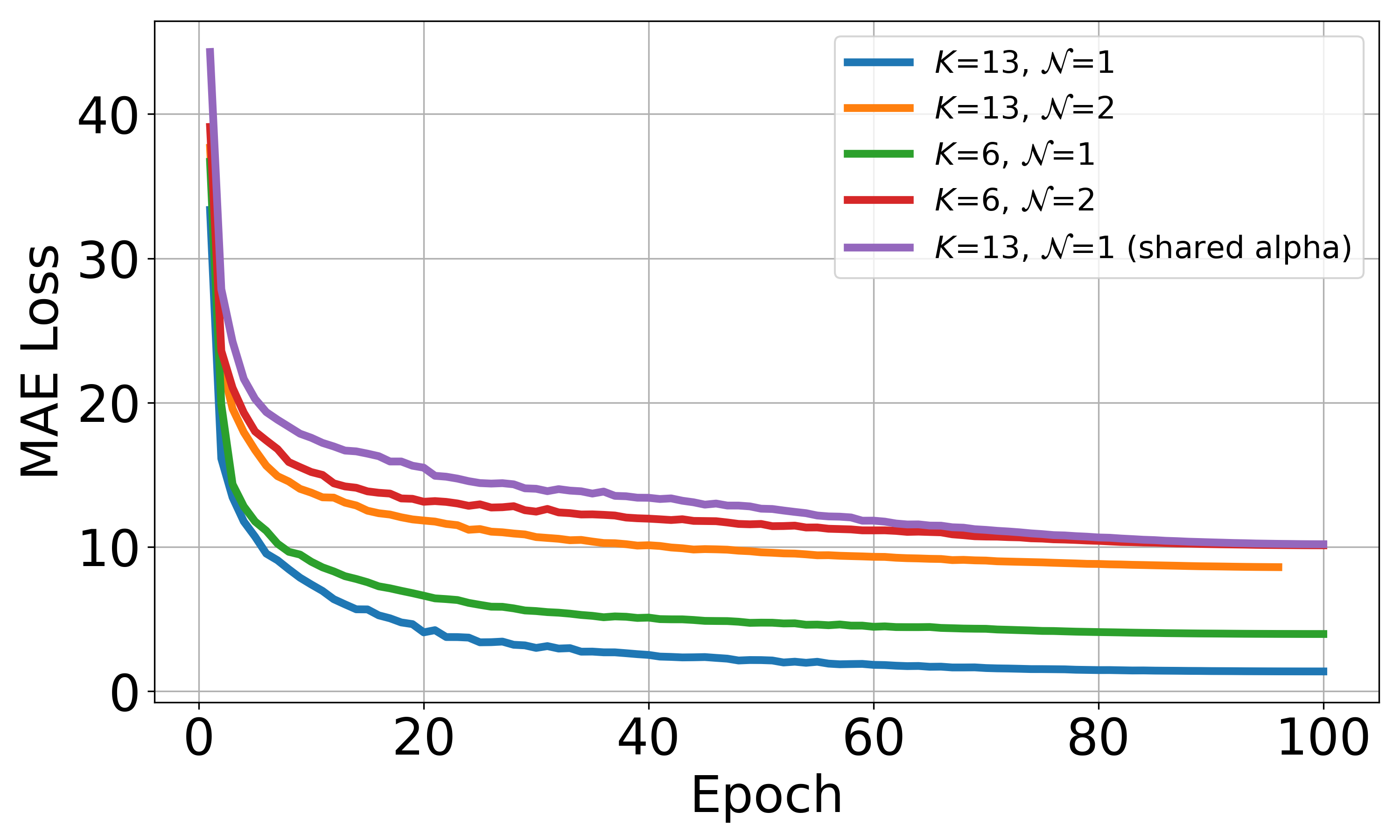}
    \vspace{-1mm}
    \label{fig:model_training}
   \end{subfigure}
   \vspace{-3mm}
   \caption{Trend of bandwidth constraint $\phi(\alpha_k)$~(left) and training loss~(right) with the change in hyperparameters ($K$ and $\mathcal{N}$).}
   \vspace{-3mm}
   \label{fig3}
\end{figure}

\vspace{-1mm}
We benchmark our model against state-of-the-art methods: Historical Last (HL)~\cite{liang2021revisiting}, LSTM~\cite{Hochreiter1997long}, D$^2$STAGNN~\cite{shao2022decoupled}, ASTGCN~\cite{guo2019attention}, PatchSTG~\cite{fang2024efficient}, VMGCN~\cite{ahmad2024variational}, CA-VMGCN~\cite{ahmad2025robust}, random projection mixer (RPMixer)~\cite{yeh2024rpmixer}, and regularized adaptive graph learning (RAGL)~\cite{wu2025regularized} . As shown in Table~\ref{tab:comparison}, our approach consistently outperforms~(in terms of accuracy) these baselines across all horizons (3/6/12 steps), for both short-term predictions ($\le 1$ hour, horizons $\le 4$) and long-term forecasts. UVMD achieves perfect signal reconstruction with no information loss, while trainable $\alpha_k$ parameters adaptively cover the full frequency spectrum. Higher $\alpha_k$ reduces mode bandwidth (lower increases it), with low values causing spectral overlap between adjacent modes. In Table~\ref{tab:hyperparameter}, we analyze three key cases: (I) $\alpha$ versus $\alpha_k$ impact, (II) window size effects, and (III) hyperparameter sensitivity ($K$ and $\mathcal{N}$). Case I demonstrates the superiority of having different $\alpha_k$ as compared to shared $\alpha$, as the Wiener-filter kernel $\frac{1}{1+2\phi(\alpha_k)(\omega-\omega_k)^2}$ in~\eqref{eq:learnable_mode_eq} prevents spectral overlap and mode merging. Our experiments use $\alpha_k$ with 35,040-length windows. Although, the shorter windows accelerate UVMD training but degrade ASTGCN performance, particularly for low-frequency modes (see Fig.~\ref{fig2}). Case III reveals $K=13$ as optimal: $K=3$ causes under-decomposition (insufficient modes) while $K=30$ causes over-decomposition (redundant features) (see Fig.~\ref{fig3}). Fig.~\ref{fig3} confirms that increasing $K$ and $\mathcal{N}$ reduces ASTGCN loss, though higher $\mathcal{N}$ creates redundant center frequencies. The shared $\alpha$ (purple) shows slower convergence and validates the benefits of having mode-specific $\alpha_k$. \looseness=-1 

\subsection{Computational Complexity} 
\vspace{-1mm}
Deep unfolding transforms iterative optimization required for VMD into a fixed-depth, differentiable network that learns data-adaptive update rules from real data~\cite{monga2021algorithm}. This approach avoids convergence loops, captures optimal descent trajectories in fewer steps, and enables faster inference with reduced computational overhead. The time complexity of original VMD, as noted earlier is, $\mathcal{O}(N\mathcal{N}KT)$, which scales with iteration count $\mathcal{N}$. Increasing $\mathcal{N}$ substantially impacts VMD computation, while UVMD maintains efficient training for $\mathcal{N}=1,2$ (higher $\mathcal{N}$ causes overfitting). Consequently, UVMD reduces decomposition times from 
267mins to 98.63s for LargeST data. \looseness=-1

\section{Conclusion}
\label{sec:conclusion}
\vspace{-1mm}
We have presented MAGN, a novel deep-unfolded framework that overcomes computational bottlenecks and spectral limitations in decomposition-based spatiotemporal forecasting. We have proposed the first unfolded VMD (UVMD) implementation for efficient decomposition of the signal and introduced adaptive mode-specific bandwidth constraints ($\alpha_k$) that automatically tune to spatial heterogeneity. Our comprehensive evaluation on the LargeST benchmark (6,902 sensors, 241 million observations) demonstrates MAGN's capabilities: achieving a 250$\times$ speedup compared to conventional variational mode decomposition while delivering 85-95\% reduction in prediction error (MAE/MAPE/RMSE) over state-of-the-art baselines and maintaining interpretable decomposition of traffic dynamics. In practice, MAGN can decompose city-scale traffic data in under a minute to enable real-time deployment in intelligent transportation systems. For future work, we propose to extend UVMD to multivariate forecasting (e.g., joint traffic flow/speed prediction), adapt the framework to other decomposition paradigms, and use MAGN for spatiotemporal forecasting problems in different applications.

\bibliographystyle{IEEEbib}
\bibliography{reference}

\end{document}